\documentclass[letterpaper, 10 pt, conference]{ieeeconf}  

    \IEEEoverridecommandlockouts                              
    
    \usepackage{multicol}
    \usepackage[bookmarks=true]{hyperref}
    \usepackage{graphicx}
    \usepackage{caption}
    \usepackage{subcaption}
    \usepackage{wrapfig}
    \usepackage{csvsimple}
    \usepackage{pgfplotstable,filecontents}
    \usepackage{algorithm}
    \usepackage[noend]{algpseudocode}
    \usepackage{lmodern}
    
    \title{\LARGE \bf
    Workspace Aware Online Grasp Planning
    }
    
    \author{Iretiayo Akinola, Jacob Varley, Boyuan Chen, and Peter K. Allen 
    \thanks{ 
            This work was supported in part by National Science Foundation grants IIS-1208153 and IIS-1527747. Authors are with the Department of Computer Science, 
            Columbia University, New York, NY 10027, USA.
            {\tt\small  (iakinola, jvarley, bchen, allen)@cs.columbia.edu},
    }}
    
\begin{document}
    
    \maketitle
    
    \graphicspath{{images}}

    
    
    
    \begin{abstract}
        This work provides a framework for a workspace aware online grasp planner. This framework greatly improves the performance of standard online grasp planning algorithms by incorporating a notion of reachability into the online grasp planning process.  Offline, a database of hundreds of thousands of unique end-effector poses were queried for feasability.  At runtime, our grasp planner uses this database to bias the hand towards reachable end-effector configurations. 
        The bias keeps the grasp planner in accessible regions of the planning scene so that the resulting grasps are tailored to the situation at hand. This results in a higher percentage of reachable grasps, a higher percentage of successful grasp executions, and a reduced planning time. We also present experimental results using simulated and real environments.
    \end{abstract}
    
    \section{Introduction}

    Grasp planning and motion planning are two fundamental problems in the research of intelligent robotic manipulation systems.  These problems are not new, and many reasonable solutions have been developed to approach them individually.  Most of the research has treated these problems as distinct research areas focusing either on: 1) how to get a set of high quality candidate grasps \cite{ciocarlie2007dimensionality}\cite{hang2014hierarchical} or 2) how to generate viable trajectories to bring the gripper to the desired hand configuration \cite{sucan2012open}\cite{schulman2014motion}\cite{diankov2008openrave}.
    While it is often convenient to treat grasp and motion planning as distinct problems, we demonstrate that solving them jointly can improve performance and reduce computation time.  Grasp planners unaware of the robot workspace and without a notion of reachability often spend significant time and resources evaluating grasps that may simply be unreachable. For example, a reachability unaware planner is equally likely to return impossible grasps that assume the robot will approach the object from the object's side furthest from the robot.  Our planner avoids unreachable areas of the workspace dramatically reducing the size of the search space. This lowers online planning time, and improves the quality of the planned grasps as more time is spent refining quality grasps in reachable portions of the workspace, rather than planning grasps that may be stable but are unreachable. 
    
    \begin{figure}
    \begin{center}
        \includegraphics[width=.475\textwidth]{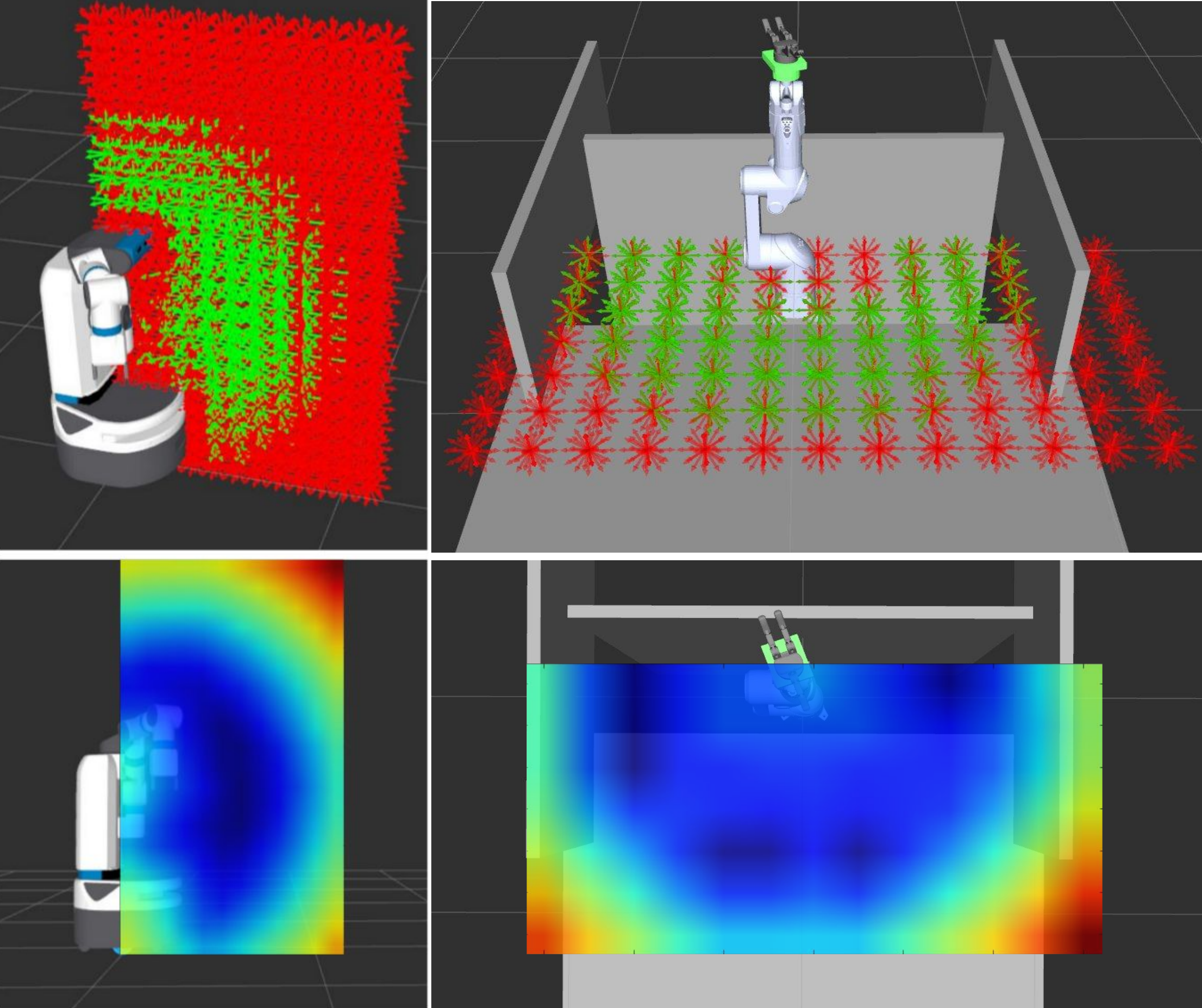}
    \end{center}
    \caption{\small Top Row: Visualization of cross sections of the precomputed reachable space for a Fetch Robot and Staubli Arm with Barrett Hand. Green arrows represent reachable poses, red arrows unreachable.  This space is computed offline, once for a given robot. Bottom Row: Signed Distance Field generated from the above reachability spaces.}
    \label{fig:front_page_figure}
    \vspace{-3mm}
    \end{figure}
    
    \begin{figure*}[t]
    \vspace{2mm}
    \begin{center}
        \includegraphics[width=1\textwidth]{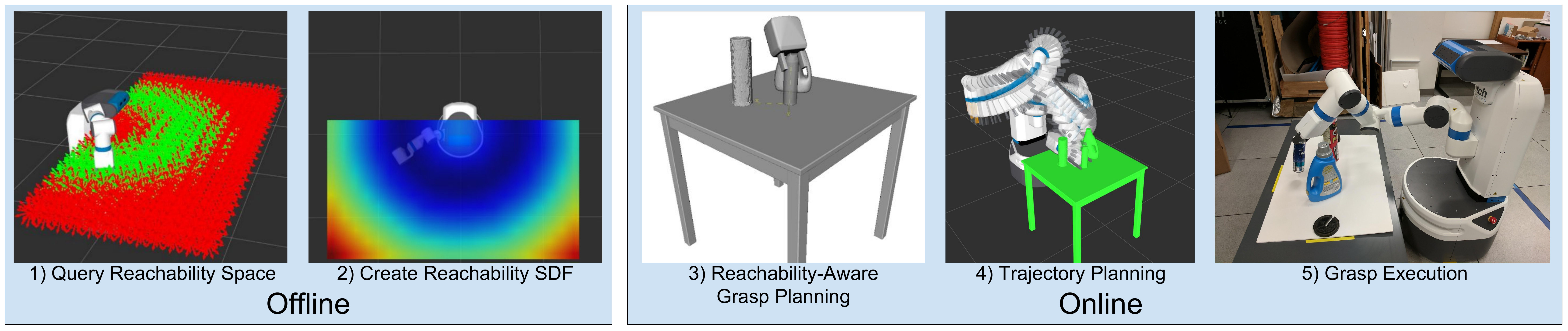}
    \end{center}
    \caption{\small Workspace Aware Online Grasping Framework - \textbf{Offline}: 1)  the robot's reachability space is queried for IK solutions that are free of collisions with the robot itself and static objects such as walls and tables. 2) An SDF is created from the reachability space. \textbf{Online}: 3) Grasp planning is quickly accomplished utilizing the reachability space SDF. 4) A motion plan for one of the planned grasps. 5) Trajectory  executed by the robot for a stable grasp. }
    \label{fig:full_pipeline}
    \vspace{-3mm}
    \end{figure*}
    
    In order to generate grasps which are stable and reachable, we propose a new energy function for use with simulated annealing grasp planners \cite{ciocarlie2007dimensionality}\cite{hang2014hierarchical}. This energy function is a weighted combination of a grasp stability term, and a grasp reachability term. Our grasp reachability term is inspired by the idea of a precomputed reachability space which has demonstrated utility in other robotics tasks \cite{porges2014reachability}\cite{vahrenkamp2009humanoid}. In our work, we generate a densely sampled reachability space for our robot as shown in Fig. \ref{fig:front_page_figure}(Top Row). This offline computation checks whether an Inverse Kinematic (IK) solution exists for a given pose.  This database in and of itself has utility for filtering lists of precomputed-grasps and quickly removing unreachable grasps, leaving only reasonable grasps as possible results. We further postprocess our reachability space and compute a Signed Distance Field (SDF) representation in Fig. \ref{fig:front_page_figure}(Bottom Row). The SDF is used in our online grasp planning energy function to guide the hand towards reachable regions of the robot's workspace. Obstacles present in the grasping scene can be embedded into the space prior to SDF generation which results in a field that repels the grasp search away from poses that collide with the obstacles.
    
    Experiments in both simulation and physical environments have been performed to assess the performance of our method on multiple objects and under various poses. We demonstrate that our reachability aware grasp planner results in a larger number of reachable grasps, a larger number of successful grasp executions, and a reduced planning time compared to other online grasp planning methods.
    
    Our method for workspace aware online grasp planning is overiewed in Fig. \ref{fig:full_pipeline}.  This framework has a number of advantages. First, it combines two closely coupled processes into one; rather than going back and forth between grasp planning and trajectory planning until a feasible grasp is found, we solve for a fully reachable grasp at once. It increases the probability of finding feasible grasps quickly since the optimization process is guided by our energy function within reachable spaces. Also, our method is well suited for online planning which in general is more adaptable and robust than the traditional offline grasp planning approach and works in the case of novel objects for which precomputed grasps do not exist.
    
    In summary, we take a closer look at the problem of grasp quality and reachability analysis of grasps in an integrated and structured manner. The contributions of this work include: 1) A novel representation of reachability space that uses signed-distance field (SDF) to provide a gradient field during grasp planning; 2) An online grasp planning system with an integrated notion of reachability; 3) A formulation that ranks a list of grasps (e.g. from a database) based on their combined reachability and grasp quality; 4) A method of embedding of obstacles into the reachability space during grasp planning;
    5) experimental results using our framework compared to other grasp planning solutions evaluated in multiple grasping scenarios.
    
    \section{Related Work}
    
\begin{figure}
    \vspace{2mm}
        \begin{center}
            \includegraphics[width=0.85\linewidth]{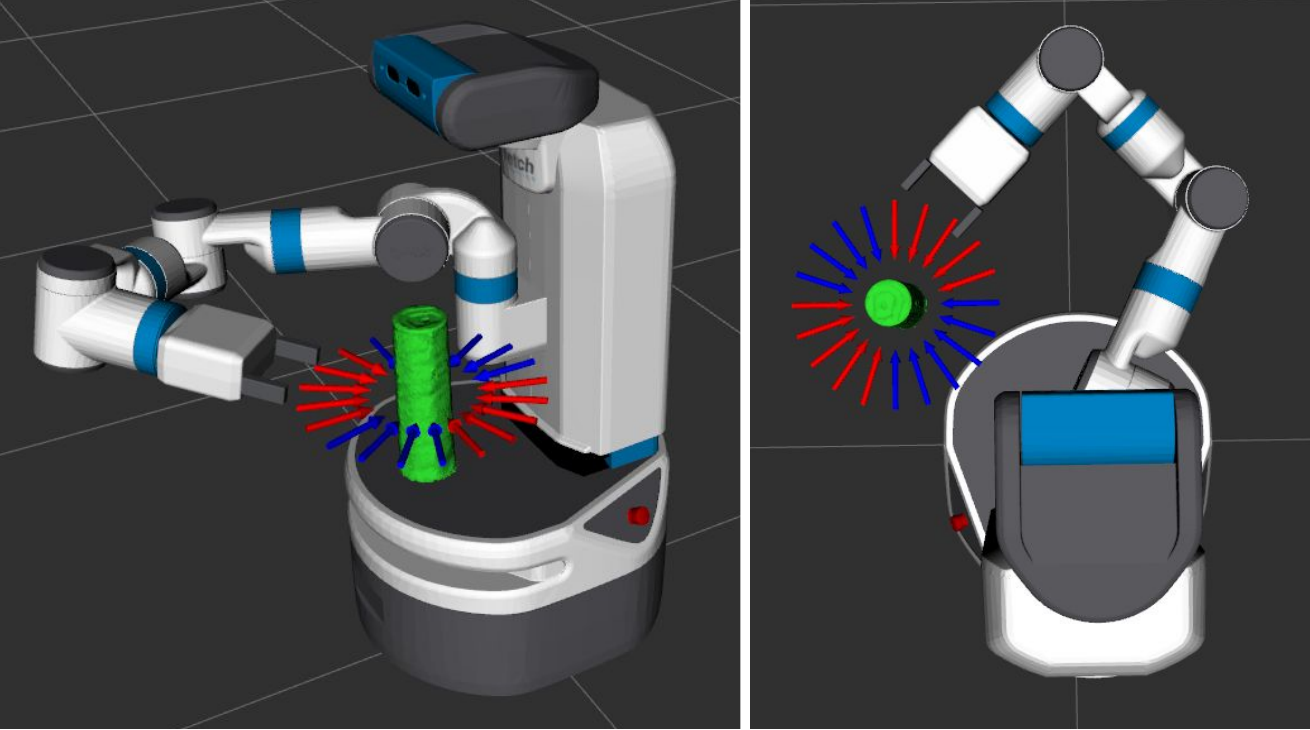}
        \end{center}
        \caption{\small Reachability results for uniformly sampled grasps (Red arrows reachable, blue unreachable).  Even for objects well within the bounds of the robot workspace, there are many invalid approach directions which are not easily modeled by simple heuristics.}
        \label{fig:around_sampled_grasps}
        \vspace{-3mm}
    \end{figure}

    Different works have highlighted the problem of generating non-reachable planned grasps from grasp planning systems. For example, a 2016 review of the Amazon Picking Challenge (APC)  \cite{correll2016analysis} explicitly highlighted the fact that many teams opted against the planning paradigm (choosing an online feedback paradigm instead). Since grasp planning metrics typically ignore the notion of reachability, a large percentage of the planned grasps end up being unreachable and not useful for the robot as illustrated in Fig. \ref{fig:around_sampled_grasps} . Hence, an extra computational step is required to check each of the planned grasps for reachability; a top performing APC team \cite{hernandez2016team} uses this post-check solution which shows that this limited partial solution is built into even competitive systems. This post-check solution has a number of drawbacks: first, the post-processing check for reachability might not contain any reachable grasp which implies that the entire grasp planning process has to be repeated. Another closely related problem with post-reachability checking is that more time is incurred when replanning grasps that fail the reachability test stage. Also, many systems, such as those in APC, are highly tuned to specific use-cases which do not generalize well especially when the grasping environment changes in small ways. Our system presents a way to automatically incorporate the environment into grasp planning by augmenting the reachability space.
    
    Many grasp planning platforms such as GraspIt!\cite{miller2004graspit}, OpenRave\cite{diankov2008openrave}, and OpenGrasp\cite{leon2010opengrasp} only consider properties specific to the end-effector during grasp planning. Given the hand description and the object, these platforms compute grasp quality for different hand-object configurations using metrics such as force/form closure and return the best quality grasps found. This approach ignores reachability of the resulting grasp as often the arm is not even modeled in the simulator, and leaves the burden of verifying the grasp outputs to the subsequent stages of robotics applications. Our work instead incorporates the notion of reachability during planning thereby increasing the quality of the output of the grasp planning processing, from the viewpoint of the subsequent stages. The reachability space aptly captures the information about the robot geometry and kinematics; our approach incorporates this information into the grasp planning process in a way that produces feasible reachable results as input for trajectory planning. Since this involves a constant time lookup, incorporating it gives extra advantage at slight computational cost.
    
    
    The reachability space presents an intuitive way to introduce soft constraints to the grasp planning process. One can restrict the allowable grasp configuration space by adjusting the reachability space to reflect such preferences. Though we have not shown it in this work, this idea has been demonstrated in a previous work \cite{berenson2007grasp}. In their work, they use an \textit{"Environment Clearance Score"} to guide grasps away from obstacles in the robot workspace. The same work introduced the idea of \textit{"Grasp-scoring function"} which uses an approximate kinematics of the robot to score a grasp. This is different from our work in the sense that we build the actual reachable space using the exact kinematics of the robot. Also, while they use the scoring function on pre-computed grasps, we use the reachability value directly for online grasp planning.
    
    The concept of offline reachability analysis exists in the literature \cite{vahrenkamp2009humanoid} \cite{porges2014reachability}. However, many of these works use offline precomputed grasps. While our approach shares the concept of offline reachability analysis with these works, we differ in that we incorporate the reachability information directly into online grasp planning. For example, a method \cite{zacharias2009online} uses capability maps and an additional inverse kinematics check to filter out sampled grasp poses. Since their grasp sampling is uniform, the chances of sampling reachable poses do not improve over time. Conversely, our method uses a gradient field to guide the grasp planning process to reachable portions. 
    Note that our work treats grasping as a self-contained task that comes up with useful outputs for subsequent stages of robotic operations while these works give more consideration to the manipulation tasks.
    
    Haustein etal.\cite{haustein2017integrating} has recently looked at combining grasp planning and motion planning. Similar to \cite{fontanals2014integrated}, their method uses a bidirectional search approach to randomly sample grasp candidates in one direction and sample arm trajectories in the opposite direction until a connection is made to form a hand grasp/arm trajectory pair. Our method is complementary to theirs in that we guide the very high-dimensional grasp goal sampling process to regions that are reachable leading to faster connection of sample grasps to the sampled arm trajectories. Our work provides a simple and effective upgrade to common standalone grasp planning approaches as well as to the inner grasp search modules of integrated grasp and motion planning systems.
    While our work is similar to \cite{berenson2007grasp} \cite{vahrenkamp2009humanoid} in being workspace aware, ours differs in a number of ways: (1) we focus on solving online grasp planning, (2) we compute the actual reachability map, not an approximation as in  \cite{berenson2007grasp}, (3) we use a novel representation of the reachability space, and (4) we use a different optimization technique, simulated annealing.

    \section{Offline Reachability Space Generation}
    A grasp is reachable if a motion plan can be found to move the arm from its current configuration to a goal configuration that places the hand at desired grasp location. This is not always possible for a number of reasons typically because no inverse kinematics(IK) solution exists to place the hand at the desired grasp pose, self collision with other parts of the robot, or collision with obstacles in the planning scene.
    \noindent\begin{minipage}[t]{0.475\textwidth}
    \vspace{-5mm}
    \begin{algorithm}[H]
    \caption{Reachability Space Generation}
    \label{alg:SDFGeneration}
    \begin{algorithmic}[1]
    \Procedure{GenerateReachabilitySpace}{}
    \State poses = uniformSampleWorkspace()
    \State pose2Reachable = $\{\}$
    \For{pose in poses}
        \State reachable = hasCollisionFreeIK(pose)
        \State pose2Reachable[pose] = reachable
    \EndFor
    \State SDF = computeSDF(pose2Reachable)
    \State \Return SDF
    \EndProcedure
    \end{algorithmic}
    \end{algorithm}
    \vspace{-2mm}
    \end{minipage}

    \subsection{Reachability Space Representation}
    We observe that this definition of the original reachability space is binary, either 1 (reachable), or 0 (not reachable), and has no gradient to indicating the direction from non-reachable regions to reachable portions of the workspace. In the context of Simulated Annealing for grasp planning, it is beneficial to provide an energy function that has a landscape that can guide solutions to more desired regions of the annealing space. This observation informed a novel signed-distance-field (SDF) representation of the reachability space that is amenable to optimization formulations. Our SDF maps a pose to a value representing the distance to the manifold or boundary between the unreachable poses and reachable poses and can be interpreted as follows $d_{sdf}=SDF(pose)$: 
    \begin{itemize}
    \item $d_{sdf} = 0$: pose lies exactly on boundary between reachable and unreachable grasp poses.
    \item $d_{sdf} > 0$: pose lies within the reachable region of the workspace, and is distance $d_{sdf}$ away from the boundary. 
    \item $d_{sdf} < 0$: pose lies outside the reachable region of the workspace, and is distance $d_{sdf}$ away from the boundary. 
    \end{itemize}
    This field can then serve as criteria for classifying a grasp as either lying in a reachable space or not.  And nearby hand configurations can be ordered based on $d_{sdf}$, their distance from this boundary.

    Grid resolution and metric choices are two important parameters for the reachability space and SDF generation. The 6D hand pose of a given grasp has the 3D translational (linear) component and the 3D rotation (angular) component, two physically different quantities with different units. We had to systematically determine good resolution levels and a good ratio between unit linear measurements and unit angular measurements, to obtain a unified 6D metric space suitable for our purpose. Effectively, our metric can be defined as:
    $$ d_{sdf} = \sqrt{||\Delta xyz/res_{lin}||^2 + r ||\Delta rpy/res_{rot}||^2  } $$
    where $\Delta xyz$ and $res_{lin}$ (in centimeters) are the translational distance and resolution respectively, $\Delta xyz$ and $res_{lin}$ (in radians) are the rotational distance and resolution, and $r$ is the relative metric ratio i.e. a distance of $res_{lin}$cm $ \equiv r* res_{rot}$rad.
    
    We varied translational resolution $res_{lin} = 5, 10, 20$cm, angular resolution $ res_{rot} = \pi/8, \pi/4, \pi/2$rad, and translational to rotational metric ratio $r =0.1,1,2,5,10$; and checked the performance of the SDF generated for each combination. First, we randomly generated 10,000 grasp poses and checked for the existence of an IK solution i.e. the grasp's reachability. Then, for each of the resolution levels, we evaluated the quality of the resulting SDF by checking what percentage of the random grasp poses were correctly classified as reachable (or not) by the SDF manifold. Reachable grasps evaluate to positive sdf values, unreachable grasps evaluate to negative sdf values. The time to generate SDF from binary reachability space was recorded in each case. From this parameter sweep, the densest resolution level $res_{lin}=5$cm,  $ res_{rot}=\pi/8$rad gave 0.992 \& 0.973 as accuracy and precision respectively but took 177secs to generate the SDF.  We observed that 10cm translational, $\pi/4$ rotational resolution levels and metric ratio $r =1$ still gave a high accuracy (0.979) and precision (0.92) with the sdf generation time under 3secs. These choices defined the metric space for the 6D hand pose reachability space: a 10cm translation and a $\pi/4$rad rotation are equidistance in the reachability space.
    Note that we take into consideration the cyclical nature of the rotational degrees of freedom during SDF generation as angles wrap around at $2\pi$.

    \subsection{Reachability Space Generation}
    Reachable-space generation is a one-time process for a given robot. Algorithm \ref{alg:SDFGeneration} details the process for generating the Reachable Space. We search a discretization of the 6D pose space of the robot to determine which configurations will be reachable by the hand. For the Fetch Robot, we sampled the workspace using the values in meters centered at the robot's base shown in Table \ref{tab:sdf_space_sampling}. In total 675,840 unique poses where queried for reachability. 102,692 were reachable and 573,148 were unreachable. This was computed using Moveit!'s IK solver which checks for a valid collision free IK solution for a given pose. The 6D pose space containing the binary IK query results was converted into an SDF using the fast marching method from scikit-fmm\footnote{https://github.com/scikit-fmm/scikit-fmm}. Once this computation is done, the SDF can be queried with any pose inside the space to determine both its binary reachability and distance to the boundary separating reachable and non reachable poses.  Reachability space generation codes and data will be made available on the project website: \url{http://crlab.cs.columbia.edu/reachabilityawaregrasping/}
    
    \begin{table}[]
    \vspace{2mm}
    \centering
    \caption{Discretization of robot workspace for SDF generation for the Fetch robot. Values are in meters (x, y, z) and radians (roll, pitch, yaw). The origin of the space is centered at the robot base. The roll was sampled less because the Fetch robot can twist its wrist to achieve any roll for a give hand pose.}
    \label{tab:sdf_space_sampling}
    \begin{tabular}{|c|c|c|l|l|l|l|}
    \hline
         & X & Y & Z & Roll & Pitch & Yaw \\ \hline
    min  & 0.0 & -1.1 & 0.0 & -PI    & -PI     & -PI   \\ \hline
    max  & 1.2 & 1.1 & 2.0 &  PI  & PI    & PI  \\ \hline
    step & 0.1 & 0.1 & 0.1 & PI   & PI/4.0    & PI/4.0   \\ \hline
    \end{tabular}
    \vspace{-2mm}
    \end{table}

    
    \section{Online Reachability-Aware Grasp Planning}
    \label{sec:GraspPlanning}
    Grasp planning is the process of finding "good grasps" for an object that can be executed using an articulated robotic end effector. A grasp is typically classified as "good" based on how well it can withstand external disturbances without dropping the object. The Ferrari-Canny method \cite{ferrari1992planning} is a common way to evaluate robot grasps; it defines how to quantitatively measure the space of disturbance wrenches that can be resisted by a given grasp. Other techniques are typically built around this metric as shown in this review \cite{roa2015grasp}. Grasp planning frameworks such as GraspIt! use these grasp metrics of force and form closure as objective functions for optimization. Since this formulation has no notion of reachability, the grasp results, while stable, might require the end effector to be placed in a pose that is impossible to reach given the robot's current location.
    
    \subsection{Simulated Annealing for Grasp Planning}
    Grasp planning can be thought of as finding low energy configurations within the hand object space. This search is done in a multidimensional space where grasps are sampled and evaluated. The 6 dimensional (x, y, z, roll, pitch, yaw) space we generated with our SDF is a subspace of the 6 + N dimensional grasp search space. The additional N dimensions represent the EigenGrasps or eigen vectors of the end effector Degree of Freedom (DOF) space as described in \cite{ciocarlie2007dimensionality}. For the Fetch (1 DOF), N=1, and for the Barrett Hand (4 DOF), N=2. These 6 + N dimensions describe both the pose and DOF values of the end effector.  The goal of grasp planning is to find low energy points in this 6 + N dimensional space.  These points represent the pose and hand configuration of "good grasps".  
    
    Simulated annealing (SA) \cite{ingber1989very} -- a probabilistic technique for approximating global optimums for functions lends itself nicely for our formulation. SA presents a number of advantages in grasp planning\cite{allen2014grasp}. For example, it does not need an analytical gradient which can be computationally infeasible; it handles the high nonlinearity of grasp quality functions; and it is highly adaptable to constraints. Our approach implicitly guides the annealing process ensuring that sampling is done in regions of reachable space which increases the probability of getting useful grasp solutions. While online grasping is typically avoided due to time cost of exploring a larger search space, this work makes online grasping more feasible since the majority of the space which is unreachable need not be searched. With our new energy formulation, the annealing process will quickly drive the hand towards reachable grasp locations. Section \ref{sec:experimental} below shows that this new energy function increases the success probability of online grasping significantly.
    
    \noindent\begin{minipage}[t]{0.475\textwidth}
    \vspace{-6mm}
    \begin{algorithm}[H]
    \caption{Contact And Potential Energy (SA-CP)}
    \label{alg:cp_energy_formulation}
    \begin{algorithmic}[1]
    \Procedure{ContactAndPotentialEnergy}{}
    \State $e_{p}$ = potentialEnergy()\label{alg_eps_cp}
    \State stable = $e_{p}$ $<$ 0
    \If{stable}
        \State \Return $e_{p}$
    \Else{}
        \State $e_{contact}$ = contactEnergy()\label{alg_contact_cp}
        \State \Return $e_{contact}$
    \EndIf
    \EndProcedure
    \end{algorithmic}
    \end{algorithm}
    \vspace{-3mm}
    \begin{algorithm}[H]
    \caption{Reachability-Aware Energy (SA-OURS)}
    \label{alg:ours_energy_formulation}
    \begin{algorithmic}[1]
    \Procedure{ReachabilityAwareEnergy}{}
    \State $e_{p}$ = potentialEnergy()\label{alg_eps_ours}
    \State $e_{reach}$ = reachabilityEnergy()\label{alg_reach_ours}
    \State stable = $e_{p}$ $>$ 0
    \State reachable = $e_{reach}$ $<$ 0
    \If{reachable \&\& stable}
        \State \Return $e_{p}$ + $\alpha_1$ $\cdot$ $e_{reach}$
    \ElsIf{reachable \&\& !stable}
        \State $e_{contact}$ = contactEnergy()\label{alg_contact_ours_v1}
        \State \Return $e_{contact}$ + $\alpha_2$ $\cdot$ $e_{reach}$
    \Else{} //!reachable
        \State $e_{contact}$ = contactEnergy()\label{alg_contact_ours_v2}
        \State \Return $e_{contact}$ + $\alpha_3$ $\cdot$ $e_{reach}$
    \EndIf
    \EndProcedure
    \end{algorithmic}
    \end{algorithm}
    \vspace{1mm}
    \end{minipage}
    
    \vspace{-1mm}
    \subsection{Novel Grasp Energy Formulation}
    \label{sec:energy_formulation}
    
    In our work, we augment the \textit{Contact and Potential} grasp energy function from \cite{ciocarlie2007dimensionality}. The \textit{Contact and Potential} energy function is described in Algorithm \ref{alg:cp_energy_formulation} and consists of a potential energy term ($e_{p}$) measuring the grasps potential to resist external forces and torques, and a contact energy term ($e_{contact}$) that defines the proximity of the hand to the object being grasped. These two terms define the grasp fitness function used by the optimization algorithm (simulated annealing) to search for good grasp configurations.
    Our \textit{Reachability Aware} method augments the \textit{Contact and Potential} energy function with a \textit{reachability} energy term ($e_{reach}$) that encodes the kinematic constraints of the robot. We use the reachability SDF ($e_{reach}:=d_{sdf}$) described previously as a regularizing term to obtain a new cost functions that drives the grasp planning optimization towards reachable hand configurations. As shown in \cite{oleynikova2016signed} discretized SDFs still model and behave like a continuous function so it can be used during the simulated annealing optimization process to drive the search towards reachable hand configurations. Our reachability-aware energy function $E$ is given as:
    \begin{equation}  \label{eq:energy_combination}
    E = G + \alpha R
    \end{equation}
    
    \noindent where $G$ is a metric of grasp quality (e.g. force closure), $R$ is a measure of reachability of a given grasp pose and $\alpha$ defines the tradeoff between the conventional grasp metric and the reachability value. To optimize the overall grasp energy, we use the simulated annealing search according to \cite{ciocarlie2009hand}.
    
    In terms of implementation, we build on the GraspIt! simulator and combine the existing energy metrics with the new reachable energy term. The energy components have forms that we exploit. The grasp energy term (as described in \cite{ciocarlie2007dimensionality} and Algorithm \ref{alg:cp_energy_formulation}) is negative when the grasp has force closure and positive otherwise. The reachability energy has similar meaning which gives four quadrants of possibilities. We chose the weights of each quadrants so the range of values for the reachability term and grasp term are of the same order of magnitude. This provides a gradient from bad to good quality (reachable and force-closure) grasps. Algorithm. \ref{alg:ours_energy_formulation} shows specifically how the energy terms are combined. $\alpha$ values were set to $\alpha_{1} = -0.1$, $\alpha_{2} = -10$, $\alpha_{3} = -10$ to obtain a function where more negative energy values correspond to more reachable stable grasps, and positive high energy values correspond to unreachable and unstable grasps.
    
    Since our reachability space is represented as a grid of discretized SDFs, we use multi-linear interpolation to get ($e_{reach}:=d_{sdf}$) for a given grasp pose from the reachability SDF grid.  Each query pose during the optimization is situated in the robot's reachable space to get the cell location and the reachability value is given as the weighted sum of the values for the $2^N$ ($N=6$) corners of the hypervoxel where the query point falls in the pre-computed SO(6) reachable space. The $2^N$ corner values are looked up in the pre-computed reachability space.
    \begin{equation}  \label{eq:multi_linear_interpolation}
    R[p_{query}] = \sum _{i=1} ^{2^N} w_i R[p_i]
    \end{equation}
    \noindent where $p_{query} =[x_q,y_q,z_q,r_q,p_q,y_q]$ is the query pose, $p_i =[x_i,y_i,z_i,r_i,p_i,y_i]$ for $i=1...2^N$ correspond to the neighbouring corners in the reachability space grid and $w_i$ is the proportion of the grid occupied by creating a volume from connecting the query point and the corner $p_i$. See \cite{wagner2008multi} for more details.

    \subsection{Embedding Obstacles in the Reachability Space}
    \label{sec:obstacle_embedding}
    
    Obstacles whose poses are not expected to change in relation to the robot can be placed in the scene while generating the initial reachability space. This approach is applicable to modeling tables, walls and fixtures around fixed base robotic arms, or static room environments for mobile manipulators. This is useful for the Staubli Arm and attached Barrett Hand as shown in Fig. \ref{fig:front_page_figure}.  This robot is fixed to the table, next to the walls, making it reasonable to incorporate these obstacles during the initial reachability space creation.
    
    In order to incorporate non-static obstacles into our precomputed reachability space, we mask out regions in the robot's binary reachability space that overlap with an obstacle prior to generating the SDF representation; this corresponds to modifying the reachable space in the first block of Fig. \ref{fig:full_pipeline}.
    Given the obstacles' geometries and poses, we first collate grid locations of the reachability space that overlap with obstacle geometries. All grasp poses at these locations (including those that were otherwise reachable) are marked as unreachable.
    Then, we regenerate an SDF representation of the resulting binary reachable space. The SDF generation is fast, taking 2-3 seconds on average, and this step is only required when the workspace changes. While this fast obstacle-masking-procedure considers only the hand (not full arm configuration) and might miss out on some other grasp poses that become infeasible due to obstacles, we've found that it is a reasonable approximation that works well in practice. The masked out obstacle locations imposes a negative field that pushes the SDF manifold further away from the obstacles and our experiments show that this effect results in an improved grasp planner.
    
    \section{Experiments}
    \label{sec:experimental}
    
    We describe different experiments in both simulation and physical environments to assess the performance of our method on multiple objects and under various poses. All experiments (simulated and real) were run on two robot platforms: the Fetch mobile manipulator and the Barrett hand mounted on a Staubli Arm (Staubli-Barrett).
    The Staubli-Barrett set-up (see Fig \ref{fig:front_page_figure}) has additional walls to limit the workspace of the robot as a safety precaution to ensure the robot does not elbow objects outside of it's workspace during grasping. This safety box heavily constrains the range of grasping directions as many grasp poses will be invalidated because of collision with the walls. Static objects such as the walls were included in the scene when constructing the reachability space offline.
    
    For each experiment, we compared a typical grasp planning method with our reachability-aware method:\\
    \noindent\textbf{Sim. Ann. Contact and Potential (SA-C\&P)}: Simulated Annealing in GraspIt! using the Contact and Potential energy function from Algorithm \ref{alg:cp_energy_formulation}.\\
    \textbf{Sim. Ann. Reachability-Aware (SA-Ours)}: Simulated Annealing in GraspIt! using the newly proposed energy function from Algorithm \ref{alg:ours_energy_formulation}.\\
    \vspace{-2mm}
    \subsection{Evaluation Metrics}
    The metrics used for evaluating our methods include:\\
    \textbf{Percent Reachable Grasps}: 
    The number of reachable grasps divided by the total number of grasps generated from a run of a given grasp planner. \\
    \textbf{Number of Required Plan Attempts}: The grasp planner returns a list of grasps ordered according to quality. We record how many of these grasps we have to go through until a valid path can be planned.\\
    \textbf{Lift Success}: Here we execute the first valid grasp from the set returned by the planner. A grasp is successful if the gripper placed at the grasp pose can successful lift the object off the ground. We use the  Klamp't simulator \cite{hauser2016robust} to test this in simulation. For the real world experiments, we executed the grasps on the real robot and checked if the object was picked up.

    \subsection{Grasp Planning with Dynamic Obstacles Experiment}
    Here we set-up a typical grasp planning scenario: in a virtual scene, three objects were placed on a table. One is the target object to be grasped while the other two are obstacles that the robot must not collide with during grasping (Fig \ref{fig:table_setup}). We used 4 different meshes as the target object. We placed the target object in 9 difficult-to-reach poses that were either at the extent of the robot's workspace, or extremely close to the robot. Both of these situations drastically limit the number of valid approach directions that the robot can use to grasp the object.
    For each pose, GraspIt!'s Simulated Annealing planner was run for varying number of planning steps using both the \textbf{SA-C\&P} and \textbf{SA-Ours} energies. We used two versions of our reachability-aware planner (\textbf{SA-Ours}), one with \textit{unmodified SDF} (Section \ref{sec:energy_formulation}) and the other with \textit{obstacles-embedded SDF} (Section \ref{sec:obstacle_embedding}). For each planner, we checked the reachability of all planned grasps and evaluate the fraction of planned grasps that are reachable after running the planner for a given number of steps. 
    
    \begin{figure}[h]
    \vspace{2mm}
    \centering
        \begin{subfigure}[h]{0.75\linewidth}
    		\centering
    		\includegraphics[width=1\textwidth,keepaspectratio]{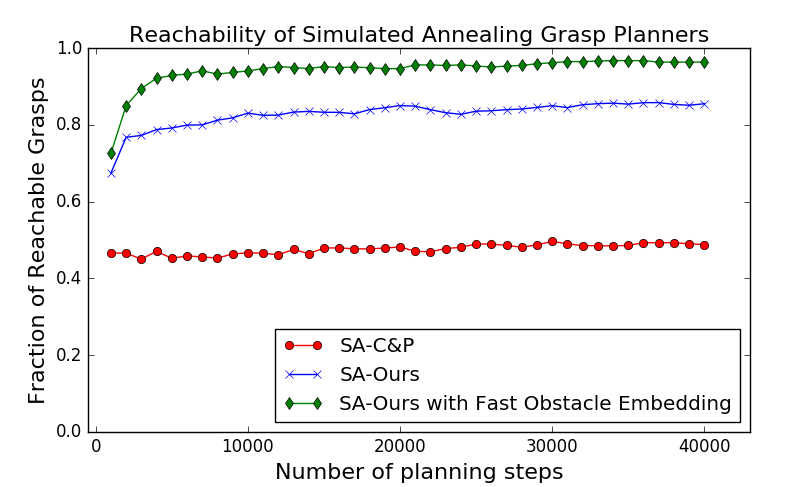}
    		\caption{\small Fetch Robot Results} 
    		\label{fig:fetch_reachability_vs_planning_steps_v1}
    	\end{subfigure}
    	\begin{subfigure}[h]{0.75\linewidth}
    		\centering
    		\includegraphics[width=1\textwidth,keepaspectratio]{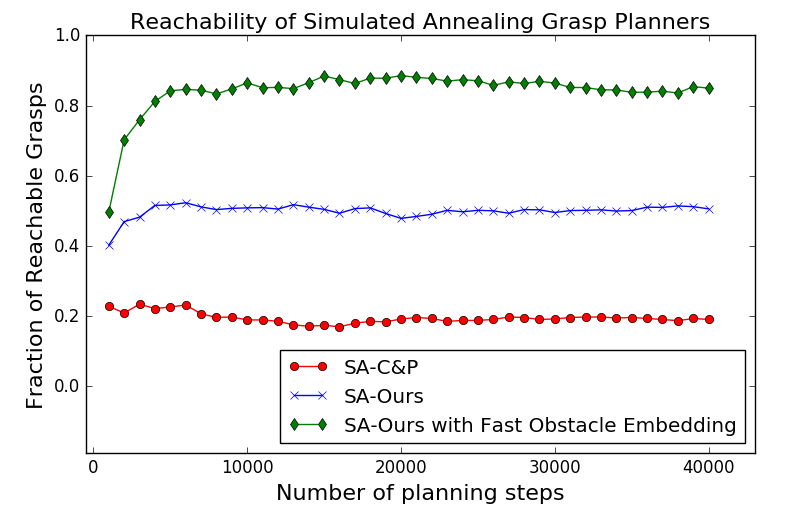}
    		\caption{\small Staubli-Barrett Results} 
    		\label{fig:barrett_reachability_vs_planning_steps_v1}	
    	\end{subfigure}
    \caption{\small Mean fraction of reachable grasps using different energy functions for varying planning duration. The red plot shows that using pure grasp planning with no notion of reachability gives grasp results that have a low chance of being reachable (48.2\% for Fetch and 18.8\% for the Staubli-Barrett). The blue plot shows that our reachability aware grasp planner results in a higher fraction ($>25\%$ increase) of reachable grasps for both robots. The green plot shows the additional gain obtained when we embed obstacles into the SDF reachability space. This results in reachable grasps that have feasible IK results and do not collide with the obstacles hence an increase in the overall fraction of reachable grasps.} 	
    \label{fig:reachability_vs_steps}
    \vspace{-3mm}
    \end{figure} 
    
    The results (Fig \ref{fig:reachability_vs_steps}) show the significance of our reachability-aware grasp planning approach compared with a typical grasp planner on the same setup. Given the same amount of time, the planners each return twenty grasp solutions and we check what fraction of them are achievable by the robot and report the average over all 36 simulated object poses (4 objects, 9 poses). As expected, a typical grasp planner that has no notion of reachability (shown in red) will yield a large percentage of grasps that are not feasible for the robot. Even with additional planning time, there is no improvement in the odds of returning reachable grasps. On the other hand, our method (shown in blue) yields significantly higher fraction of reachable grasps ($>25\%$ improvement). In addition, the fraction of reachable grasps increases quickly with planning time as our method optimizes for both stability and reachability of grasps.
    While the blue plot uses the original offline SDF, the green plot shows that by embedding the new obstacles into the precomputed SDF, we are able to increase the percentage of reachable grasps (Fetch: 95.7\%, Staubli-Barrett: 86.1\%) compared to (84.3\%, 50.2\%) respectively when the precomputed SDF only avoids self collision and static objects such as the walls.
    
    After demonstrating that our method increases the feasibility of grasp planning results, we also observe that this method also leads to a significant speedup of the grasp planning process since the reachability energy term guides the annealing process quickly to reachable regions hence reducing the search space. Once GraspIt! returns a list of grasps, we iterate through the list in order of grasp quality and attempt to plan a valid path for each grasps. When a valid grasp is found, we use the Klampt! simulator to get the lift success.
    For a simple parallel jaw gripper like the Fetch gripper, there was no significant difference in the percentage of lift success (\textbf{SA-C\&P}: \textbf{0.95}  and \textbf{SA-Ours}: 0.93) but our method requires less number of motion planning attempts (\textbf{SA-C\&P}: 2.75 and \textbf{SA-Ours}: \textbf{1.10}) to find a valid grasp in the list and overall returns a higher percentage of useful grasps (reachable and lift success). Our grasp energy formulation ensures that a top ranked grasp has a high chance of being reachable. The difference in grasp quality was more pronounced with the 4 DOF Barrett hand. Our method not only requires less number of motion planning attempts (\textbf{SA-C\&P}: 6.2  and \textbf{SA-Ours}: \textbf{1.27}), it also achieves 20\% higher lift success rate (\textbf{SA-C\&P}: 0.75 and \textbf{SA-Ours}: \textbf{0.95}). This is because our planner spends most of it's time refining grasps in the much-reduced reachable regions.

    \subsection{Real Robot Crowded Scene Experiment}
    
    \begin{figure}
    \vspace{2mm}
    \centering
        \begin{subfigure}[h]{0.8\linewidth}
    		\centering
    		\includegraphics[width=1\textwidth,keepaspectratio]{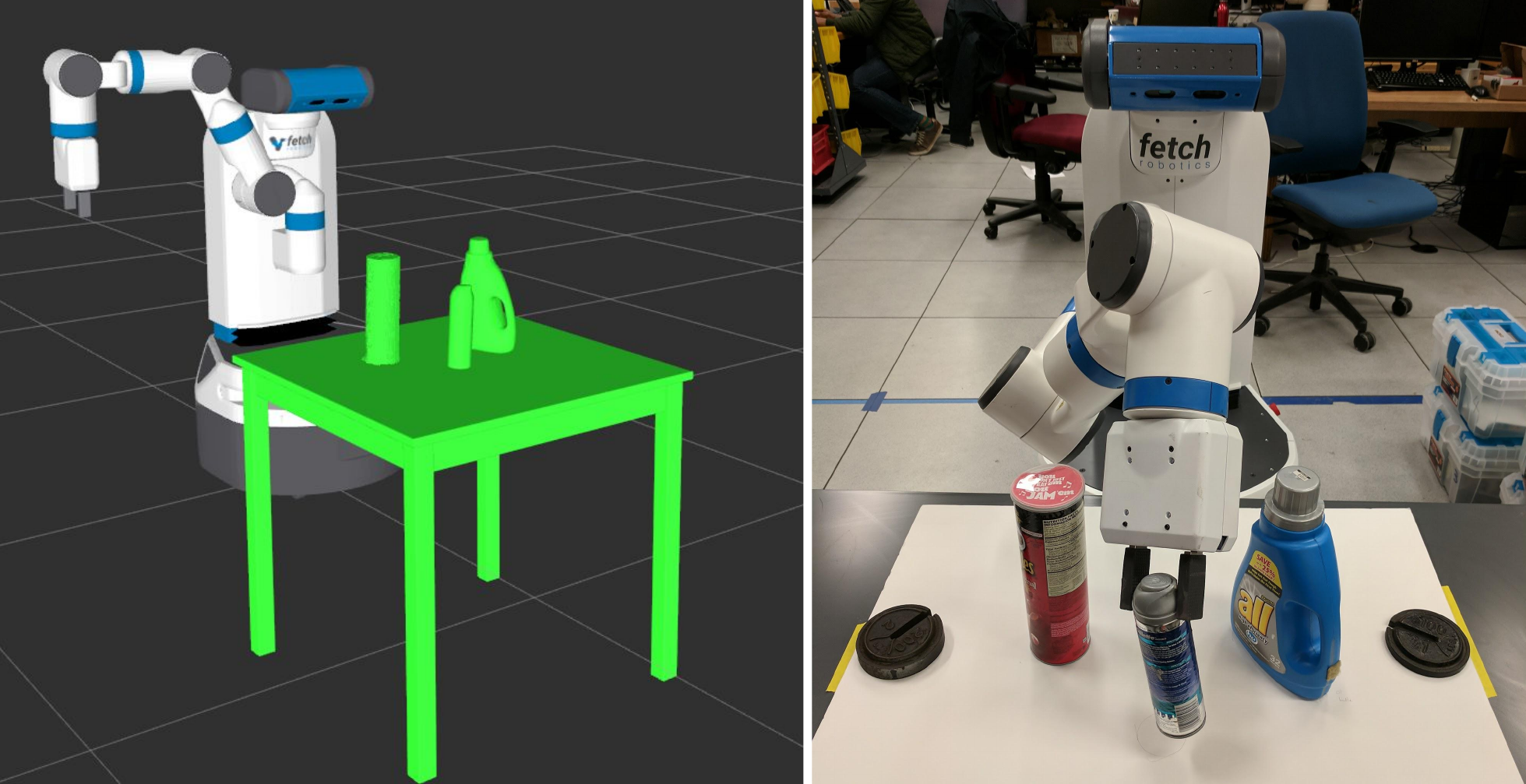}
    		\label{fig:fetch_table_setup}
    	\end{subfigure}
    	\begin{subfigure}[h]{0.8\linewidth}
    		\centering
    		\includegraphics[width=1\textwidth,keepaspectratio]{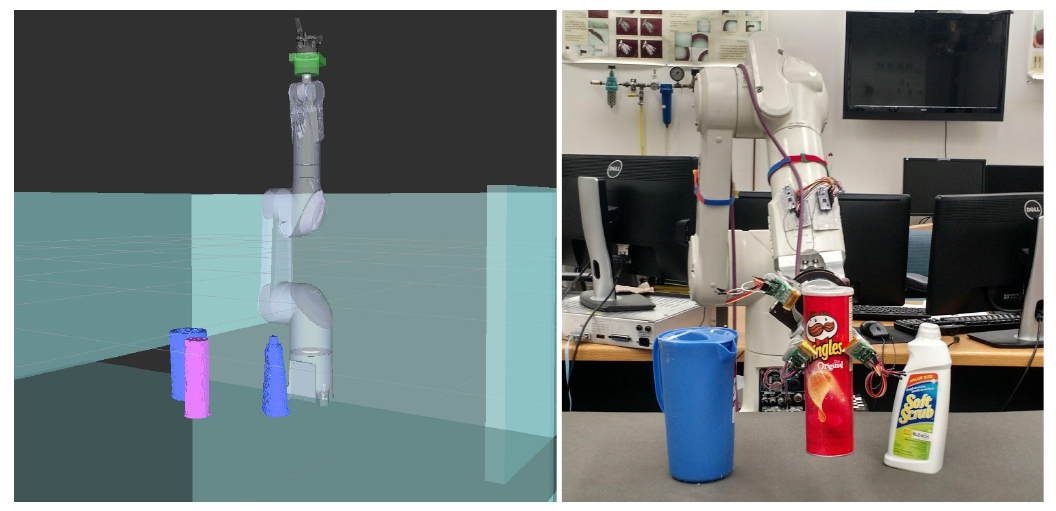}
    		\label{fig:barrett_workspace_setup}	
    	\end{subfigure}
    	
        \vspace{-2mm}
        \caption{\small Crowded scene for real world experiments. \textbf{Top} (Fetch Robot): Our Reachability-Aware planner (SA-Ours) was able to successfully grasp the shaving cream bottle 3/3 times, while annealing without the reachability space (SA-C\&P) failed 2/3 times. \textbf{Bottom} (Staubli-Barrett Robot): SA-Ours successfully grasped the pringles bottle 5/5 times, while SA-C\&P failed 3/5 times.}
        \label{fig:table_setup}
    \end{figure}

    To verify our planner and demonstrate that it works outside of simulation, the simulated annealing based grasp planner was run with a variable number of annealing steps and the success rate was reported in Table \ref{tab:live_results} and \ref{tab:live_results_barrett}. This experiment compares two grasp energy formulations: \textbf{SA-Ours} and \textbf{SA-C\&P}.
    
    \begin{table}[]
    \centering
    \caption{Grasp success results on real robot (Fetch) with a crowded scene. Each method was given 3 attempts to plan and execute a grasp on the shaving cream bottle.}
    \label{tab:live_results}
    \begin{tabular}{|l|c|c|c|}
    \hline
    \multicolumn{4}{|c|}{Crowded Scene Grasp Planning (Fetch)} \\ \hline
    \begin{tabular}[c]{@{}c@{}}Search \\ Energy\end{tabular}  & \# steps &
    \begin{tabular}[c]{@{}c@{}}Success \\ Rate\end{tabular}   & \begin{tabular}[c]{@{}c@{}}Mean Grasp\\Planning Time (s)\end{tabular} \\ \hline
    SA-Ours &10K     & \textbf{100.0\%}   &      8.706      \\ \hline
    SA-C\&P &10K  & 33.3\% &       8.360    \\ \hline
    SA-C\&P &40K & 66.6\% &     32.31      \\ \hline
    \end{tabular}
    \vspace{-3mm}
    \end{table}
    
    \begin{table}[]
    \vspace{2mm}
    \centering
    \caption{Grasp success results on real Staubli-Barrett robot with a crowded scene (Fig \ref{fig:table_setup}). Each method was given 5 attempts to plan and execute a grasp on the pringles bottle.}
    \label{tab:live_results_barrett}
    \begin{tabular}{|l|c|c|c|}
    \hline
    \multicolumn{4}{|c|}{Crowded Scene Grasp Planning (Barrett)} \\ \hline
    \begin{tabular}[c]{@{}c@{}}Search \\ Energy\end{tabular}  & \# steps &
    \begin{tabular}[c]{@{}c@{}}Success \\ Rate\end{tabular}   & \begin{tabular}[c]{@{}c@{}}Mean Grasp\\Planning Time (s)\end{tabular} \\ \hline
    SA-Ours &10K     & \textbf{100.0\%}   &      11.36      \\ \hline
    SA-C\&P &10K  & 40\% &       9.53    \\ \hline
    SA-C\&P &40K & 60\% &     31.98      \\ \hline
    \end{tabular}
    \vspace{-3mm}
    \end{table}
    
    Multiple objects were placed in the planning scene as shown in Fig \ref{fig:table_setup}. The additional objects reduce the range of possible grasps since the obstacles make many grasp poses infeasible. Note that both methods are aware of the obstacles during grasp planning and avoid grasps that put the hand in collision with obstacles. Table \ref{tab:live_results} shows that our method, \textbf{SA-Ours}, is able to achieve grasp success -- with the Fetch robot picking the object three times out of three within the limited planning time. Conversely, \textbf{SA-C\&P} fails to produce a reachable grasp 2/3 times when run for 10,000 steps, and fails once even when it is allowed to run for 40,000 steps.  Despite being allowed to run for a long duration, the naive planner spends much of its time exploring the back half of the bottle which is completely unreachable to the Fetch robot.
    
    We repeated the same experiment with Staubli-Barrett robot (bottom of Fig \ref{fig:table_setup}). Though the objects were placed roughly at the centre of the workspace, the presence of walls and distractor objects significantly limit the range of feasible grasps for the target object. Each method had 5 trials and the results followed the same pattern: \textbf{SA-Ours} achieves grasp success all five times within the limited planning time (10,000 steps) while \textbf{SA-C\&P} fails 2/5 times even when run for 40,000 steps. 
    \vspace{-1mm}

    \section{Conclusion and Future Work}
    
    This work provides a framework for a workspace aware grasp planner. This planner has greatly improved performance over standard online grasp planning algorithms because of its ability to incorporate a notion of reachability into the online grasp planning process.  This improvement is accomplished by leveraging a large precomputed database of over 675,840 unique end-effector poses which have been tested for reachability.  At runtime, our grasp planner uses this database to bias the hand towards reachable end effector configurations. This bias allows the grasp planner to generate grasps where a significantly higher percentage of grasps are reachable, a higher percentage result in successful grasp executions, and the planning time required is reduced. It has been experimentally tested in both simulated and physical environments with different arm/gripper combinations.
    
    Several future research directions include: utilizing our computed reachability workspace to help mobile robots navigate to optimal locations for manipulating objects and improvements for speeding up the process of incorporating dynamic objects into our notion of reachability to further assist the grasp planner.
    \bibliographystyle{IEEEtran}

\begin{thebibliography}{10}
\providecommand{\url}[1]{#1}
\csname url@samestyle\endcsname
\providecommand{\newblock}{\relax}
\providecommand{\bibinfo}[2]{#2}
\providecommand{\BIBentrySTDinterwordspacing}{\spaceskip=0pt\relax}
\providecommand{\BIBentryALTinterwordstretchfactor}{4}
\providecommand{\BIBentryALTinterwordspacing}{\spaceskip=\fontdimen2\font plus
\BIBentryALTinterwordstretchfactor\fontdimen3\font minus
  \fontdimen4\font\relax}
\providecommand{\BIBforeignlanguage}[2]{{%
\expandafter\ifx\csname l@#1\endcsname\relax
\typeout{** WARNING: IEEEtran.bst: No hyphenation pattern has been}%
\typeout{** loaded for the language `#1'. Using the pattern for}%
\typeout{** the default language instead.}%
\else
\language=\csname l@#1\endcsname
\fi
#2}}
\providecommand{\BIBdecl}{\relax}
\BIBdecl

\bibitem{ciocarlie2007dimensionality}
M.~Ciocarlie, C.~Goldfeder, and P.~Allen, ``Dimensionality reduction for
  hand-independent dexterous robotic grasping,'' in \emph{Intelligent Robots \&
  Systems}.\hskip 1em plus 0.5em minus 0.4em\relax IROS, 2007, pp. 3270--3275.

\bibitem{hang2014hierarchical}
K.~Hang, J.~A. Stork, and D.~Kragic, ``Hierarchical fingertip space for
  multi-fingered precision grasping,'' in \emph{Intelligent Robots and Systems
  (IROS)}.\hskip 1em plus 0.5em minus 0.4em\relax IEEE, 2014, pp. 1641--1648.

\bibitem{sucan2012open}
I.~A. Sucan, M.~Moll, and L.~E. Kavraki, ``The open motion planning library,''
  \emph{IEEE Robotics \& Automation Magazine}, vol.~19, no.~4, pp. 72--82,
  2012.

\bibitem{schulman2014motion}
J.~Schulman, Y.~Duan, J.~Ho, A.~Lee, I.~Awwal, H.~Bradlow, J.~Pan, S.~Patil,
  K.~Goldberg, and P.~Abbeel, ``Motion planning with sequential convex
  optimization and convex collision checking,'' \emph{IJRR}, vol.~33, no.~9,
  pp. 1251--1270, 2014.

\bibitem{diankov2008openrave}
R.~Diankov and J.~Kuffner, ``Openrave: A planning architecture for autonomous
  robotics,'' \emph{Robotics Institute, Pittsburgh, PA, Tech. Rep.
  CMU-RI-TR-08-34}, vol.~79, 2008.

\bibitem{porges2014reachability}
O.~Porges, T.~Stouraitis, C.~Borst, and M.~A. Roa, ``Reachability and
  capability analysis for manipulation tasks,'' in \emph{ROBOT2013: First
  Iberian Robotics Conference}, pp. 703--718.

\bibitem{vahrenkamp2009humanoid}
N.~Vahrenkamp, D.~Berenson, T.~Asfour, J.~Kuffner, and R.~Dillmann, ``Humanoid
  motion planning for dual-arm manipulation and re-grasping tasks,'' in
  \emph{IROS}.\hskip 1em plus 0.5em minus 0.4em\relax IEEE, 2009.

\bibitem{correll2016analysis}
N.~Correll, K.~E. Bekris, D.~Berenson, O.~Brock, A.~Causo, K.~Hauser, K.~Okada,
  A.~Rodriguez, J.~M. Romano, and P.~R. Wurman, ``Analysis and observations
  from the first amazon picking challenge,'' \emph{IEEE Transactions on
  Automation Science and Engineering}, 2016.

\bibitem{hernandez2016team}
C.~Hernandez, M.~Bharatheesha, W.~Ko, H.~Gaiser, J.~Tan, K.~van Deurzen,
  M.~de~Vries, B.~Van~Mil, J.~van Egmond, R.~Burger \emph{et~al.}, ``Team
  delft's robot winner of the amazon picking challenge 2016,'' \emph{preprint
  arXiv:1610.05514}, 2016.

\bibitem{miller2004graspit}
A.~T. Miller and P.~K. Allen, ``Graspit! a versatile simulator for robotic
  grasping,'' \emph{IEEE Robotics \& Automation Magazine}, vol.~11, no.~4, pp.
  110--122, 2004.

\bibitem{leon2010opengrasp}
B.~Le{\'o}n, S.~Ulbrich, R.~Diankov, G.~Puche, M.~Przybylski, A.~Morales,
  T.~Asfour, S.~Moisio, J.~Bohg, J.~Kuffner \emph{et~al.}, ``Opengrasp: a
  toolkit for robot grasping simulation,'' in \emph{International Conference on
  Simulation, Modeling, and Programming for Autonomous Robots}, 2010, pp.
  109--120.

\bibitem{berenson2007grasp}
D.~Berenson, R.~Diankov, K.~Nishiwaki, S.~Kagami, and J.~Kuffner, ``Grasp
  planning in complex scenes,'' in \emph{Humanoid Robots, 2007 7th IEEE-RAS
  Internatn'l Conf. on}, pp. 42--48.

\bibitem{zacharias2009online}
F.~Zacharias, C.~Borst, and G.~Hirzinger, ``Online generation of reachable
  grasps for dexterous manipulation using a representation of the reachable
  workspace,'' in \emph{Advanced Robotics. ICAR 2009. International Conference
  on}.\hskip 1em plus 0.5em minus 0.4em\relax IEEE, pp. 1--8.

\bibitem{haustein2017integrating}
J.~A. Haustein, K.~Hang, and D.~Kragic, ``Integrating motion and hierarchical
  fingertip grasp planning,'' in \emph{Robotics and Automation (ICRA), 2017
  IEEE Int'l. Conf.}, pp. 3439--3446.

\bibitem{fontanals2014integrated}
J.~Fontanals, B.-A. Dang-Vu, O.~Porges, J.~Rosell, and M.~A. Roa, ``Integrated
  grasp and motion planning using independent contact regions,'' in
  \emph{Humanoid Robots, 2014 14th IEEE-RAS International Conference on}.\hskip
  1em plus 0.5em minus 0.4em\relax IEEE, 2014, pp. 887--893.

\bibitem{ferrari1992planning}
C.~Ferrari and J.~Canny, ``Planning optimal grasps,'' in \emph{Robotics and
  Automation, 1992. Proceedings., 1992 IEEE International Conference on}.\hskip
  1em plus 0.5em minus 0.4em\relax IEEE, 1992, pp. 2290--2295.

\bibitem{roa2015grasp}
M.~A. Roa and R.~Su{\'a}rez, ``Grasp quality measures: review and
  performance,'' \emph{Autonomous Robots}, vol.~38, pp. 65--88, 2015.

\bibitem{ingber1989very}
L.~Ingber, ``Very fast simulated re-annealing,'' \emph{Mathematical and
  computer modelling}, vol.~12, no.~8, pp. 967--973, 1989.

\bibitem{allen2014grasp}
P.~K. Allen, M.~Ciocarlie, and C.~Goldfeder, ``Grasp planning using low
  dimensional subspaces,'' in \emph{The Human Hand as an Inspiration for Robot
  Hand Development}, 2014, pp. 531--563.

\bibitem{oleynikova2016signed}
H.~Oleynikova, A.~Millane, Z.~Taylor, E.~Galceran, J.~Nieto, and R.~Siegwart,
  ``Signed distance fields: A natural representation for both mapping and
  planning,'' in \emph{RSS 2016 Workshop: Geometry and Beyond-Representations,
  Physics, and Scene Understanding for Robotics}.\hskip 1em plus 0.5em minus
  0.4em\relax University of Michigan, 2016.

\bibitem{ciocarlie2009hand}
M.~T. Ciocarlie and P.~K. Allen, ``Hand posture subspaces for dexterous robotic
  grasping,'' \emph{IJRR}, vol.~28, no.~7, 2009.

\bibitem{wagner2008multi}
R.~Wagner, ``Multi-linear interpolation,'' \emph{Beach Cities Robotics}, 2008.

\bibitem{hauser2016robust}
K.~Hauser, ``Robust contact generation for robot simulation with unstructured
  meshes,'' in \emph{Robotics Research}, 2016.

\end{thebibliography}


    \end{document}